\documentclass{article}

\usepackage[preprint,nonatbib]{neurips_2022} 

\usepackage[utf8]{inputenc}
\usepackage[T1]{fontenc}    

\usepackage{graphicx}

\usepackage{amssymb,amsfonts,amsmath}
\usepackage{nicefrac}       
\usepackage{microtype}      

\usepackage[style=apa]{biblatex}
\let\cite\parencite

\usepackage{hyperref}

\usepackage{xcolor}

\usepackage{booktabs}       
\usepackage{multirow}

\title{Nothing makes sense in deep learning, \\ except in the light of evolution}
\author{%
  Artem Kaznatcheev \\
  Department of Biology\\
  University of Pennsylvania\\
  Philadelphia, PA\\
  \texttt{kaznatcheev.artem@gmail.com} \\
  \And
  Konrad Paul Kording \\
  Departments of Biological Engineering \& Neuroscience \\
  University of Pennsylvania\\
  Philadelphia, PA\\
  \texttt{koerding@gmail.com} \\
}

\usepackage[colorinlistoftodos,textsize=small]{todonotes}
\usepackage{hyperref}

\usepackage{xstring}
\newcounter{parnum}

\usepackage{caption}
\newcommand{\omited}[1]{}

\begin{document}

\maketitle

\begin{abstract}
Deep Learning (DL) is a surprisingly successful branch of machine learning.
The success of DL is usually explained by focusing analysis on a particular recent algorithm and its traits.
Instead, we propose that an explanation of the success of DL must look at the population of all algorithms in the field and how they have evolved over time.
We argue that cultural evolution is a useful framework to explain the success of DL.
In analogy to biology, we use `development' to mean the process converting the pseudocode or text description of an algorithm into a fully trained model. 
This includes writing the programming code, compiling and running the program, and training the model.
If all parts of the process don't align well then the resultant model will be useless (if the code runs at all!).
This is a constraint.
A core component of evolutionary developmental biology is the concept of deconstraints -- these are modification to the developmental process that avoid complete failure by automatically accommodating changes in other components.
We suggest that many important innovations in DL, from neural networks themselves to hyperparameter optimization and AutoGrad, can be seen as developmental deconstraints. 
These deconstraints can be very helpful to both the particular algorithm in how it handles 
challenges in implementation
and the overall field of DL in how easy it is for new ideas to be generated.
We highlight how our perspective can both advance DL and lead to new insights for evolutionary biology. 
\end{abstract}

\section{Evolution of deep learning}
\label{sec:evoDL}

Deep learning (DL) allows computers to learn enough from data to recognize faces~\cite{schroff2015facenet}, play go~\cite{schrittwieser2020mastering}, and model text~\cite{brown2020language} successfully.
Coffee allows scientists to stay awake long enough to write another paper on what makes DL algorithms successful.
How is it possible that caffeine has just the right structure to keep us awake enough to write this paper?
Caffeine is an arrangement of 24 atoms in just the right way to resemble adenosine enough to occupy its receptors in brain cells but sufficiently different that it does not cause sleepiness~\cite{huang2005adenosine}. 
From the molecular perspective, this coincidence seems miraculous. 
Out of all the possible molecules made out of 24 atoms, the probability that a random one interacts with our neuroreceptors in just the right way is unbelievably small. 
This makes the proximate explanation -- in terms that reduce to the interactions between atoms and molecules -- for the success of caffeine feel unsatisfying relative to the ultimate explanation in terms of history.

To be fully satisfied by an account of caffeine, we need to look at the history of this unlikely molecule. 
This history was both a process of cultural evolution -- consuming herbs that affect us -- and before that, a process of biological evolution -- the coffee plant evolving a defense mechanism against insects that share some of their brain chemistry with us~\cite{denoeud2014coffee}. 
Caffeine shares similar structures to our neurotransmitters because we share an evolutionary history. 
Evolution explains why coffee is successful.

What matters for explanations of molecules like caffeine matters even more for explanations of whole organisms.  
This is what \textcite{D64} meant when he famously wrote that ``nothing makes sense in biology except in the light of evolution”. 
We cannot fully understand the success of a biological molecule or organism by just looking at the structure of that molecule or organism.
We need to also look at the natural history that led to the evolution of that molecule or organism and how it relates to other organisms and the environment.

In this paper we argue that the same kind of natural history is needed to understand the success of deep learning.
We see traditional explanations of the success of certain traits in DL algorithms as explaining how the trait works -- the explanation of the trait is in terms of the algorithms or the data set.
We are advocating for explaining why the trait exists -- we seek explanations of the trait in terms of how the algorithm came about from prior algorithms and how the trait evolved from earlier ancestor traits.
In other words, we cannot fully understand the success of deep learning algorithms or their rapid improvement by just looking at the structure of any given algorithm. 
We need to also look at the cultural history of the human scientific and engineering processes that created these algorithms.

In the 1960s, neural networks could not solve nonlinear problems~\cite{minsky1969perceptrons}. 
In the 1980s, they could solve very simple problems, like recognizing digits~\cite{lecun1989backpropagation}. 
In the 1990s, they became able to play backgammon at a strong level~\cite{tesauro1994td}. 
In 2012 they could recognize real-world objects~\cite{krizhevsky2012imagenet}. 
And now neural networks, rebranded as deep learning, are better than humans at recognizing objects, playing go, and transcribing spoken text~\cite{lecun2015deep}. 
It’s clear that the deep learning of today is not the neural networks of the 1960s -- the population of algorithms changed. 

\begin{figure}[ht]
    \centering
    \includegraphics[width=\textwidth]{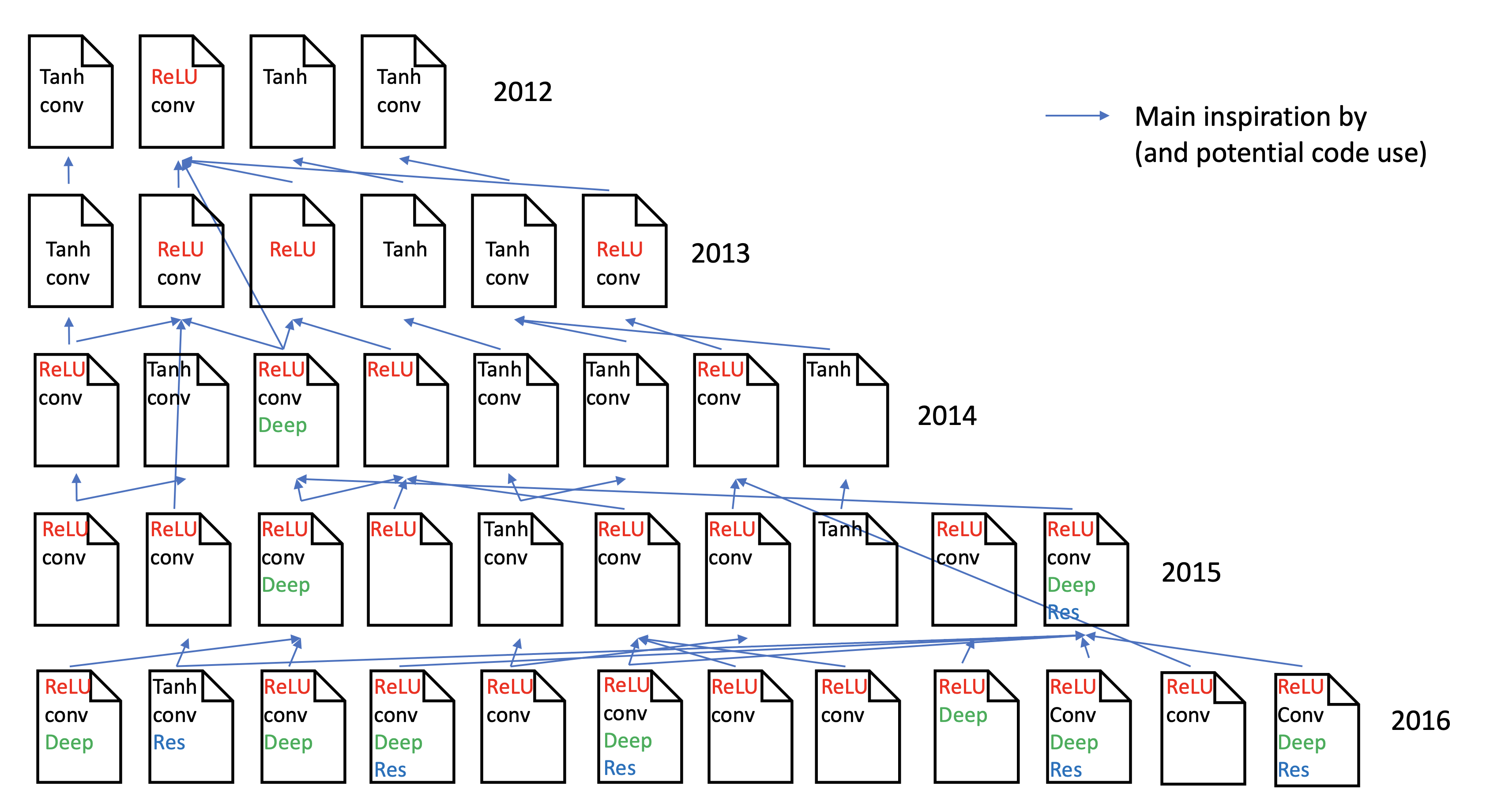}
    \caption{{\bf Heritability and selection in deep learning}. 
    We can view the papers and code published in each year of NeurIPS as a population (here: one row per year).
    In these populations we see variation in multiple traits.
    Here, on each paper, we write a few typical traits (Tanh, ReLU, conv, Deep, Res) that are used in that paper (or the algorithm it describes).
    For example: some models might use ReLU activation functions while others use Tanh, some might have many hidden layers (deep) and others few, etc.
    The presence or absence of these traits is often inspired by work published in prior years of NeurIPS (here shown with blue arrows).
    Over many years, some traits become more common (here: ReLU) while others start to disappear (here: Tanh), suggesting a process of selection.
    }
    \label{fig:gen}
\end{figure}

The theory of evolution explains why and how populations change over time. 
If we look, in Figure~\ref{fig:gen}, at each generation of DL papers (e.g. those published each year at the conferences NeurIPS or ICML), we see:

\begin{enumerate}
\item[(H1)] variation in algorithms, 
\item[(H2)] heritability of traits of algorithms between generations, and
\item[(H3)] differential survival of algorithms over time.
\end{enumerate}

In other words, we see the three hallmarks of evolution in the progression of populations of DL algorithms.
If we look at resources like PapersWithCode then we also have a rich fossil record of these past generations.
This record tracks both the dynamic of broad traits of deep learning algorithms, and their performance on various benchmarks, alongside with the source code that specifies these organisms.

The presence of these three hallmarks of evolution and the natural history of the DL field, shows us that deep learning algorithms receive two kinds of data. 
First is the traditional training data of any particular algorithm on any particular task. 
Second -- and perhaps more important -- is the information encoded in the evolutionary history of how these algorithms came about -- information on what worked and what did not work in the past~\cite{InfoLoad}. 
Cultural evolution here is a meta-algorithm for the population of DL algorithms to learn about its particular environment.

From the No-Free-Lunch Theorem~\cite{NFL97, Devroye1983-tk} we know that deep learning is not better than other algorithm over the set of \emph{all} possible problems. 
But we do not care about all possible problems. 
We care about problems in the DL field's particular environment: the world as it happens to be.
Through its history, and successive generations of NeurIPS and ICML conferences, the field has been incorporating many inductive biases that improve performance on the world as it happens to be. 
It is this evolutionary history of a successful algorithms that sets up the biases of the algorithm to align with what has been typical in the world.

Of course, this is a decisively qualitative account of why DL algorithms are successful.
Just like the early naturalists in biology, who did not know the DNA-basis of heredity, we do not yet know how exactly the three hallmarks of evolution are implemented in the DL field.
It is not as simple as the code base or the manuscript text, and requires future work to identify the hereditary basis -- i.e., the ``genes" or ``memes" -- for deep learning algorithms.
The quantitative tools of modern evolutionary biology could help us answer these questions in the future.
We give some example questions for each of the three hallmarks of evolution, each questioned numbered the same as the corresponding hallmark:
\begin{enumerate}
\item[(Q1)] How is variation in DL algorithms created between generations (See Fig. \ref{fig:EvoDevo})?
\item[(Q2)] What exactly is inherited between generations of ICML/NeurIPS papers and  can we make precise estimates of the heritability of the various traits in DL papers? 
Can we identify the relevant `genes’ that encode these traits? 
\item[(Q3)] What are the selective pressures on the population of DL algorithms created by our benchmarks and community standards?
\end{enumerate}
Answering these questions is useful to both our understanding of deep learning and as a case study in the broader field of the cultural evolution of science~\cite{WO_CS2022}.

\section{Facilitated variation in biological evolution}

The ``survival of the fittest'' (i.e., natural selection) part of evolution gives us some clues as to why deep learning algorithms work in practice but leaves us with the question: why was there such a quick and drastic improvement in their performance in recent years?
After all, we usually expect evolution to be slow and gradual. 
Existing work in the cultural evolution of science has primarily focused on heritability (H2 and Q2) and selection (H3 and Q3)~\cite{WO_CS2022}.
But deep learning gives us the opportunity to look carefully at the first hallmark (H1) and the question of the generation of variation or ``arrival of the fittest'' (Q1).
As we consider the ``arrival of the fittest'', the history of deep learning might seem quite different from biological evolution in one particular way:
new mutations in biology are random but new ideas in deep learning do not seem to be random.

Popular accounts of biological evolution often attribute all the creative aspects of evolution to selection.
In deep learning, however, we do \emph{not} attribute all creative aspects of the field to the hard working peer reviewers and program committee members, nor to our careful citation practices.
And while we read a competitor's paper, we might -- in our frustration -- imagine that it is just a random jumble of prior work. 
We know that this is not really the case.
For when we design the models in our own latest paper, we certainly feel like we are deploying creativity and not just randomly modifying and assembling what we have seen before.
In other words, unlike simple views of biological evolution, the variation that we create in a new deep learning paper does not seem random.

Does this mean that an evolutionary analysis does not apply?
No, it means that a simple model of biological evolution that equates random genetic change with random phenotypic change does not apply.
But this simple model of evolution does not apply to biology, either.

\subsection{EvoDevo background and foundations}

An organism is not simply a genotype labeled by a fitness value.
Real biological organisms are the outcome of rich developmental processes that create a phenotype from a genotype (and environment).
This phenotype then results in some fitness for the organism in some environment.
Thus, the developmental processes that constitute an organism create a genotype-phenotype-fitness map.
Due to these processes and this map, uniformly random variation at the genetic level can result in non-random variation at the phenotypic level.
To make sense of this, we have to turn to evolutionary developmental biology and the theory of facilitated variation~\cite{GK07,KG08book}.
Here we will lay out the relevant biological concepts and terminology before applying them to the history of DL in Section~\ref{sec:EvoDevoDL}.

The many different core developmental processes that make up an organism have to be robust to noise and failure in other developmental processes and the environment.
This robustness necessitates feedback and information flow between these core processes. 
There is an \emph{interchangeability of cues}, since the signals that flip the conditionals in a developmental process can be the result of components encoded by genes that are internal to the process, or the external environment, or even other developmental processes.
This interchangeability of cues 
allows for \emph{weak regulatory linkages} to be formed between processes without significant changes to the processes themselves -- the biochemical signals act as a kind of universal API for the library of modules that make up the organism.

Weak regulatory linkages allows the organism's developmental processes to signal each other and adapt to disturbances in other processes.
\emph{Somatic adaptation} and \emph{somatic accommodation} are the ability of some parts of the developing body (called `soma') to accommodate changes in other parts.
For example, the blood vessels in the human body are not all individually specified, but develop in response to signals from other cells in the body requesting oxygen, nutrients, or the removal of waste products.
This somatic adaptation can be very helpful to both (1) the individual organism in how it handles noise in the environment or errors in other processes 
and (2) the organism's lineage by how somatic adaptation handles mutations in the genome.
It is this second benefit of somatic adaptation to the evolvability of a population that interests us here.

\begin{figure}[ht]
    \centering
    \includegraphics[width=\textwidth]{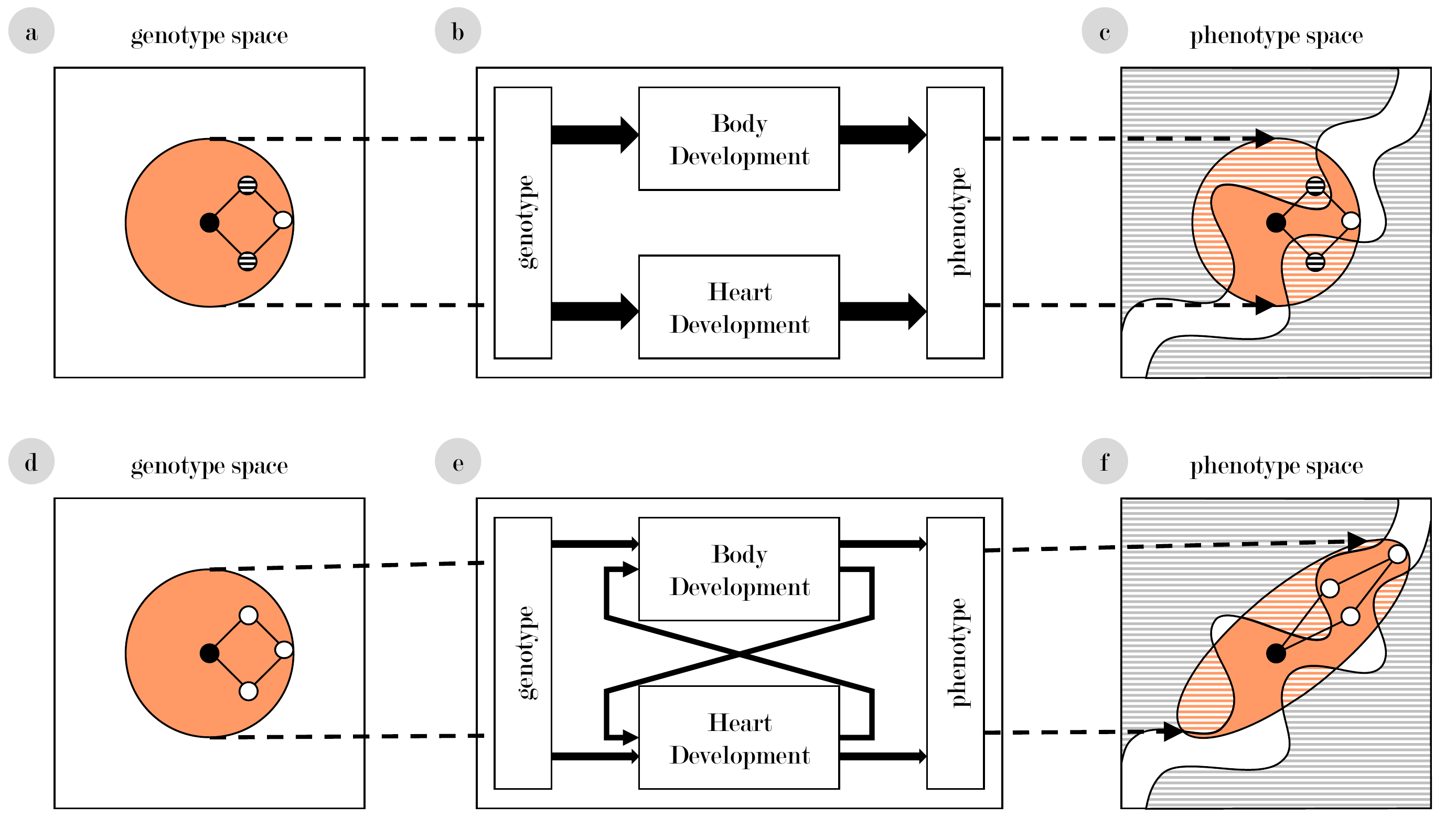}
    \caption{\textbf{Example of evolutionary deconstraint.}
    Moving from left to right, 
    we see how the genotype space \textbf{(a,d)} is transformed by 
    development \textbf{(b,e)} into 
    the phenotype space \textbf{(c,f)} and how this transforms the space of nearby mutations (orange circles) from the wildtype (black solid circle).
    Development is shown as an information flowchart \textbf{(b,e)} leading from a genotype to a phenotype through the development of two traits of the body and heart.
    Since selection happens at the level of phenotypes, we represent viable phenotypes as white regions and unviable phenotypes as the grated regions in phenotype space \textbf{(c,f)}.
    The top panels \textbf{(a,b,c)} consider the case before a developmental deconstraint is introduced, and
    the bottom panels \textbf{(d,e,f)} considers the case after a developmental deconstraint is introduced through weak regulatory linkage between body development and heart development.
    Weak regulatory linkage can be seen in \textbf{(e)} as the introduction of a feedback cycle from body development to heart development and vice-versa -- this allows these two processes to coordinate during the organism's development.
    Inset (in \textbf{(a,c,d,f)}) are small examples of fitness landscapes with two point mutations leading to a double mutation, viable mutations are shown as white circles and unviable mutations are shown as grated circles --
    this highlights how a deconstraint can replace a case of reciprocal sign epistasis \textbf{(a,c)} by a smooth landscape \textbf{(d,f)}.
    }
    \label{fig:EvoDevo}
\end{figure}

Evolution is made difficult when there are many constraints between multiple components~\cite{EvoPLS,EvoJAIR} -- called `\emph{epistasis}’ in biology.  
For an extreme example, that we visualize in Figure~\ref{fig:EvoDevo}, imagine two developmental processes that produce two traits which together form the organism's phenotype.
In our imagined environment, these two traits have to line up correctly.
If they don't line up correctly then there is complete failure of the organism -- the coordination of the traits is essential for the structure of the organism.
If the processes that produce these two traits are not signaling each other during development then this forms a constraint on the evolution of the organism:  one trait cannot be significantly modified by mutations without compensatory changes in the other interdependent trait. 
It is unlikely that a random mutation can simultaneously make just the right complimentary changes to both of the processes.

A developmental deconstraint is any process in the organism that allows a developmental process in that organism to avoid structural failure by automatically accommodating changes in other developmental processes of the organism or the environment. 
This 
converts a structural parameter that was under genetic control into a parameter that is under developmental or environmental control.
When such deconstraints are discovered by evolution, they ease evolution by allowing the previously constrained traits to vary more easily.
This can create newly accessible paths through or around fitness valleys that previously blocked evolution.
Deconstraints also allow random genetic mutations to have non-random phenotypic effects that can `line up' better with the selective pressures of the environment.
Since the genotype-phenotype-fitness map was created by prior evolution in a given enironment, this map can facilitate the partial alignment of the phenotypic effects of new genetic mutations with the demands of the environment.

It can also be helpful to think of developmental deconstraints from the perspective of how they move information between the genome, development, and environment.
Here we see a sharp contrast between \emph{genetic assimilation}~\cite{W42,W53} -- where information moves from development and learning into the genome -- and what we would like to name \emph{developmental compression} -- where information moves from the genome into development.
For a generic example of developmental compression: Suppose process A uses an internal cue X encoded in its genome -- it is possible for another process B (or even the environment) to produce cue X to affect process A.
The replacement of an internally encoded X by the products of other processes can then reduce how much information needs to be encoded in the description of process A -- so development can act as a way of compressing the heritable information.
This is most drastic when a previously internal cue can be replaced by an external environmental cue.
For a concrete example of developmental compression we can consider the brain.
As pointed out by \textcite{zador2019critique} and others, there is simply not enough space in the human genome to code for the location and synaptic connection between all the neurons in our brain.
Instead, developmental processes are specified by a compressed code that is then decoded during development -- from embryonic and fetal development to psychological learning in infancy, childhood and later life -- and build the brain in conjunction with input from other developmental processes and the environment.
This kind of interaction between evolution and development, along with considerations of information flow, is central to current evolutionary theory.

\section{EvoDevo of developmental compression in deep learning}
\label{sec:EvoDevoDL}

In biology, development is a broad concept that spans levels from embryology to social learning. 
More narrowly, in the central dogma of molecular biology, development includes every process that creates an organism based on its genetic specification.
In computer science, we will consider a general specification of a model or algorithm as the scientist-facing description -- usually as pseudocode or text.
And we will use `development' to mean every process downstream of the general specification.
For a clear example -- all processes during compilation or runtime would be under `development'. We might even consider as `development' the human process of transforming pseudocode in a paper into a programming language code.

In this section, we will consider three different levels of development in deep learning algorithms. 
We will discuss the training of neural networks with backpropagation, hyperparameter optimizers, and AutoGrad as developmental deconstraints on three distinct levels. We will describe in some detail how development at each of these three levels relates to central concepts from biology and ultimately give a list in Table~\ref{tab:DLexamples} of further deconstraints we observe in the DL field.

\subsection{Backpropagation for training weights: weak regulatory linkages}

We can define the phenotype of a classifier by its input-output behavior.
If we specify the behavior of this classifier using a crisp logical framework like a Turing Machine then even small changes to the machine or circuit will potentially produce drastic and unexpected changes to the phenotype that can easily result in the structural failure of the classifier.
For a developmental biologist, this programming approach means extremely strong and brittle regulatory control over the phenotype.
By introducing the idea of weights and activation functions, neural networks alleviate this brittleness to small changes without sacrificing expressiveness. Introducing this continuousness and differentiability to the computational model allows for weak regulatory control of the classifier's phenotype, it allows us to tune weights to obtain good performance. Neural networks themselves can be seen as a deconstraint on programming.
%

Backpropagation as a method for updating these weights based on labeled examples from training data then puts these weakly regulated processes under developmental control. If weights are the developing body or soma of the neural net then backpropagation is the somatic adaptation.
This allows scientists to automatically train the neural network on labeled training data instead of relying on the cultural evolutionary process of scientists manually trying to write better classifiers from scratch. `Learning' is what computer scientists call this developmental process of backpropagation modifying weights.
Many computational neuroscientists also equate this with learning as used in biology and psychology -- although this analogy is questioned by some~\cite{zador2019critique}. Freeing humans from tuning weights by trial and error, backpropagation is a major developmental deconstraint for the creation of classifiers.

\subsection{Hyperparameter optimization: interchangeability of cues}

We can define a middle level of development as the process that produces the higher level of development that is the learning behavior. 
We can then focus on more foundational levels of development than the learning of weights. 
In the early days of neural networks, scientists had to manually specify the various parameters of the neural network: learning rate, number of neurons, local architecture properties, etc. 
These parameters were, however, sufficiently abstracted so that they could be specified as variables which regulated the rest of the learning process.
This abstraction allowed the interchangeability of cues -- the variables could potentially be modified either by the scientist or the algorithm.

Originally these parameters or interchangeable cues were largely under ``genetic'' control and change in them was due to the cultural evolutionary process of scientists producing new models or tweaking existing ones. 
The advent of hyperparameter-optimizers put these parameters under the control of the developmental process of the algorithm. 
This took the pressure of having these numbers be optimized off of the shoulders of scientists and moved these cues to be set by the learning algorithm itself. 
What was under genetic control, could now be under the control of a developmental process that could automatically search through the settings that make these components coordinate with each other in the best way. 
Thus, we saw developmental compression: a movement from genetics (manually specifying the model's hyperparameters) into development (where the hyperparameters of the model optimize themselves). 
%

\subsection{Automatic differentiation: somatic adaptation}

We can define the lowest level of development as the developmental processes that take place before the algorithm receives any training data.
We consider automatic differentiation as an example developmental process that does not even need to use  training data.

In 2002, when Konrad built neural networks to describe unsupervised learning of neurons in brains, he had to specify how the loss function depended on the inputs and the parameters of the neurons -- 
which was not all that hard. 
But then he had a harder manual task: he had to use calculus to differentiate the loss to calculate the gradients. 
This made the overall model building hard because Konrad needed to write two parts of the code: one that calculates the loss, and one that calculates the gradients. 
Any mismatch between these two parts was likely to lead to structural failure and a non-viable model. 
This was an evolutionary constraint.

In 2022, when Artem builds a neural network with just a few lines of PyTorch, he only needs to write one code -- the code calculating the loss.
The entire second part of calculating gradients is done automatically by the developmental process of AutoGrad~\cite{baydin2018automatic}. 
The constraint that the loss calculation and the gradient need to match is thus automatically satisfied by somatic accommodation.
Artem does not even need know calculus to code a new loss function.

\subsection{Innovations in DL: toolbox of life}

Anyone who has created classifiers pre- and post- backpropagation, hyperparameter optimization, or automatic differentiation (AutoGrad) can attest to the massive increase in the ease of making new models or, as we argue here, in the ease of evolution.
The evolutionary constraint faced by Konrad's generation of scientists, no longer constrains Artem's generation of scientists.
The developmental deconstraints of AutoGrad, hyperparameter optimization, and backpropagation made innovation easier. 
It became more likely that (random) tinkering with a given project would succeed.

Above, we just talked about  three of the deconstraints that the deep learning community has discovered. 
We list a few more notable examples of biology-like ways of simplifying the creation of classifiers in Table~\ref{tab:DLexamples} -- 
these have been a remarkable feature of the evolution of the deep learning field. 

\begin{table}[ht]
    \centering
    \begin{tabular}{c|c}
        \textbf{Interdependent processes} & \textbf{Developmental deconstraint}  \\
        \hline
         Architecture vs Learning rule & Gradient descent~\cite{bottou2018optimization} \\
         Compute vs Gradient & Autograd~\cite{baydin2018automatic} \\
         Network vs Parameters & Hyperparameter optimization \\
         & \cite{claesen2015hyperparameter} \\
         Gradient complexity vs Gradient amplitude & Residual connections~\cite{he2016deep} \\
         Learning rate of one parameter vs Another & Adam~\cite{kingma2014adam} \\
         Design network vs Parallelizing across GPUs & Transformers~\cite{vaswani2017attention} \\
         Specify network vs Initialization & Principled initialization \\
         & \cite{glorot2010understanding}  \\
         Data vs Preprocessing & Batch Normalization ~\cite{ioffe2015batch} \\
         Learning rate vs Batch Size & Principled learning rates~\cite{krizhevsky2014one}
    \end{tabular}
    
    \caption{\textbf{Some deconstraints and developmental compression in the evolution of deep learning.}
    On the left are examples of interdependent processes that are coupled by some constraint and were previously `genetically-controlled' and on the right are deconstraints that brought them under developmental control with a reference to a representative paper.}
    \label{tab:DLexamples}
\end{table}

What does our reflection on deconstraints mean for the future of deep learning?
We should identify constraints between the developmental processes that we use to create classifiers and 
look for further opportunities to replace what is currently done by scientists, with processes by which the processes automatically adapt to one another. 
To the level that the deep learning field is evolutionary, we should strive towards building-in deconstraints. 
Fortunately, this is readily doable. 
We may analyze all past papers in deep learning, and ask which traits or innovations go well with one another. 
This will suggest potential constraints. 
Then we can write code that automatically facilitate the coordination of these processes. 
For example, we could analyze which modules systematically appear together in successful models and then implement libraries that automatically tune the parameters of these modules to behave well together.
This is remarkably similar to ongoing developments in the automatic machine learning field (e.g. \cite{feurer2020auto}) where strategies similar to successful past strategies are prioritized. 
Would this automate innovation in DL?


\section{Discussion}


In this paper, we have proposed an interpretation of the evolution of the DL field based on evolutionary developmental biology.
We argued that cultural evolution can help explain the success of deep learning.
Instead of focusing on explanations of how a trait of DL algorithms works we are trying to explain why the trait exists.
Instead of focusing on a particular paper or algorithm, we view papers/algorithms that come out in each year of ICML, NeurIPS, or similar venues as a population.
This lets us see the three hallmarks of evolution in the DL field: (H1) variation in algorithms, (H2) heritability of the traits of algorithms between generations, and (H3) differential survival of algorithms over time.
This means that DL algorithms receive training through two channels: (1) training data for a particular algorithm, and (2) information encoded in the evolutionary history of how these algorithms originated.
It is the second information channel that helps DL algorithms get around the No-Free-Lunch Theorem by incorporating inductive biases that improve performance on the world as it happens to be.

Each of the three hallmarks (H1, H2, H3) raises associated questions (Q1, Q2, Q3) about the evolution of DL. 
We focused on (Q1): How is variation in DL algorithms created between generations?
To explain why new innovations in DL do not feel like uniformly random mutations, we linked the process converting the pseudocode or text description of an algorithm into a trained model to the concept of `development' in biology.
The requirement that all parts of this process have to coordinate in order to produce a viable model results in a constraint on the evolution of DL.
Biological evolution has innovated `deconstraints' to get around similar kinds of requirements in biology.
Deconstraints are modifications to the developmental process that avoid complete failure of the process by automatically accommodating changes in other components.
In Section~\ref{sec:EvoDevoDL} and Table~\ref{tab:DLexamples}, we suggested that many important innovation in DL (from neural networks themselves to hyperparameter optimization to AutoGrad) can be seen as developmental deconstraints.
These deconstraints are helpful to both (1) the particular algorithm in how it handles challenges in implementation and (2) the overall field of DL in how easy it is for new ideas to be generated.

What do these deconstraints on the evolution of DL mean?
The advantage (1) to the particular algorithm helps explain why these innovations evolved.
But the advantage (2) to the field is in how the deconstraints increase evolvability.
We think that, by eliminating constraints, deconstraints make discovering new innovations easier and their impact larger.
This might be one of the reasons why we saw such a drastic increase in the speed of innovation in DL in recent years.

Why should these deconstraints on the evolution of DL matter to the NeurIPS community?
If we want our field to continue its rapid improvement then we should aim to find current constraints in the development of DL models.
Once we find the coupled developmental processes that produce these constraints, we can search for ways to automatically have the processes accommodate each other and thus remove the constraint.

By looking at DL, we are looking at only one branch of a large phylogenetic tree of ML algorithms.
It would be interesting to study other branches of ML, and science more generally, as evolutionary processes.
By viewing culture as evolving, we may ask what constraints were encountered and which past innovations in our cultures acted as deconstraints.

\subsection{Limitations and Future Work}
\label{sec:limits}
There are limitations to our evolutionary perspective on deep learning.
We think that these limitations provide great opportunities for new directions for research.

Here we have shown how to frame central ideas from the deep learning field from an evolutionary developmental perspective. 
However, we have not been quantitative. 
Like an early naturalist awed by the beauty and diversity of life, we reported our general observations on the history of deep learning.
But evolutionary biology has progressed to a highly quantitative field --
we have not used these rich quantitative methods of modern evolutionary biology.
Future work should be quantitative --
the data is all there! 
The publication record of the field is open:
the catalog of all past accepted NeurIPS paper is publicly accessible and sites like PapersWithCode already contain much of the relevant code and characterizations of phenotypes (e.g. trained model).
Regardless of the data source, the perspective we put forward here could be more meaningful with a careful quantitative analysis of the relevant traits of algorithms and the developmental processes that generate them. 
Such a quantitative account would benefit not only our understanding of DL, but it would also advance evolutionary biology.
Rich ``fossil records'' like the NeurIPS archive and PapersWithCode can allow us to `wind back the the tape of life' for DL in a way that we seldom can in the naturalistic setting of traditional biology. 
In biology, we also cannot easily study organisms that did not live.
But in DL, we can potentially look at the many papers that were rejected by NeurIPS each year rather than just the few that were published.
Since natural selection is an eliminative process, these is some sense in which it would be even more helpful to know which algorithms did not survive.

Here we focused on papers/code as the units that make up the evolving population. 
This view may be incomplete in many ways. 
For example, in an alternative view, scientists are the relevant unit: they develop during their doctoral and/or postdoctoral training and then mostly write similar papers for the rest of their career.
This is also known as the sea squirt theory of tenure~\cite{Dennett1991}: like a juvenile sea squirt, an early career academic has a rudimentary nervous system to learn and navigate the world, but once the sea squirt finds a suitable spot to cling to for life, it has no need for a brain and proceeds to digest it -- like an academic who found a permanent position.
Apart from the philosophy of science consequences of contrasting papers/code (which are much closer to `beliefs' in formal epistemology models~\cite{WO_CS2022}) to scientists as units of selection, it is a huge difference for the analogy to developmental biology.
Whereas we made development for a paper/code level of selection precise by defining it as the process converting the pseudocode or text description of an algorithm into a trained model, the scientists level of selection developmental process that converts a fresh undergraduate into an active scientists requires deeper investigates.
Which of these two units is the right level of selection in the DL field is currently unknown and both levels are likely to be relevant.
We hope that our evolutionary perspective on DL will open the door for quantitative investigation of this question.


\subsection{Potential impact on society}
\label{sec:impact}
A final aspect to consider for future work is how understanding the evolution of DL can help us improve our field.
Just as studying evolutionary medicine can be useful for a physician working on human maladies, 
studying the cultural evolution of DL might help us treat some of the systemic biases and inequalities that our work produces or reinforces.
Awareness of the evolutionary forces at play, can help us shape our field for the better.
As we noted at the end of Section~\ref{sec:evoDL}, DL algorithms receive two kinds of data: 
(1) training data for a particular algorithm, and (2) information encoded in the evolutionary history of how these algorithms originated.
We are already becoming increasingly aware of how training data can propagate or reinforce undesirable bias in our models~\cite{barocas2017fairness,MedImageBias,BP21}.
Our evolutionary perspective suggests that we should also critically examine how the historical origins of our algorithms and the community that selected them can potentially introduce undesirable bias into the whole population of learning algorithms themselves~\cite{B21,hooker2021moving}.
Given the role of the community in forging the selective pressures on the population of DL algorithms this perspective also invites us to examine not just the personal failures of individual models, papers or research teams; 
but the structural failures of the field as a whole.
For example, the original draft of this paper cited less than 8\% women -- this under-representation is a pervasive problem in our field~\cite{NeuroCiteBias}.
Being mindful of the evolution and natural history of our field has the potential to both speed up our field's progress and make future work more fair and equitable than our past.

\printbibliography

\end{document}